\documentclass{article}
\usepackage{spconf,amsmath,graphicx}
\usepackage{amsthm}
\usepackage{multirow}
\usepackage{array}
\usepackage{cite}
\usepackage{amssymb,amsfonts}
\usepackage{textcomp}
\usepackage{color}
\usepackage[utf8]{inputenc}
\usepackage[T1]{fontenc}
\usepackage{float}
\usepackage{booktabs}
\usepackage{algorithm,algorithmicx,algpseudocode}

\def\L{{\cal L}}

\def\0{{\mathbf 0}}
\def\1{{\mathbf 1}}

\def\f{{\mathbf f}}

\def\s{{\mathbf s}}

\def\u{{\mathbf u}}
\def\v{{\mathbf v}}
\def\w{{\mathbf w}}

\def\C{{\mathbf C}}
\def\D{{\mathbf D}}

\def\F{{\mathbf F}}

\def\F{{\mathbf F}}

\def\I{{\mathbf I}}

\def\L{{\mathbf L}}

\def\P{{\mathbf P}}

\def\U{{\mathbf U}}

\def\W{{\mathbf W}}
\def\X{{\mathbf X}}

\def\Tr{{\text{Tr}}}

\def\ie{{\textit{i.e.}}}
\def\eg{{\textit{e.g.}}}

\def\cE{{\mathcal E}}

\def\cG{{\mathcal G}}

\def\cL{{\mathcal L}}
\def\cM{{\mathcal M}}

\def\cV{{\mathcal V}}

\def\0{{\mathbf 0}}
\def\1{{\mathbf 1}}

\def\f{{\mathbf f}}

\def\s{{\mathbf s}}

\def\u{{\mathbf u}}
\def\v{{\mathbf v}}
\def\w{{\mathbf w}}

\def\C{{\mathbf C}}
\def\D{{\mathbf D}}

\def\F{{\mathbf F}}

\def\F{{\mathbf F}}

\def\I{{\mathbf I}}

\def\L{{\mathbf L}}

\def\P{{\mathbf P}}

\def\U{{\mathbf U}}

\def\W{{\mathbf W}}
\def\X{{\mathbf X}}

\def\Tr{{\text{Tr}}}

\def\ie{{\textit{i.e.}}}
\def\eg{{\textit{e.g.}}}

\def\cE{{\mathcal E}}

\def\cG{{\mathcal G}}

\def\cL{{\mathcal L}}
\def\cM{{\mathcal M}}

\def\cV{{\mathcal V}}

\title{Sparse Graph Learning with Spectrum Prior for Deep Graph Convolutional Networks}
\name{Jin Zeng$^{\dagger}$
\qquad Yang Liu$^{\ddagger}$
\qquad Gene Cheung$^{\star}$ 
\qquad Wei Hu$^{\ddagger}$
\thanks{Corresponding author: Wei Hu (forhuwei@pku.edu.cn).}
}

\address{$^{\dagger}$ Tongji University, Shanghai, China ~~~~~~
$^{\ddagger}$ Peking University, Beijing, China \\
$^{\star}$ York University, Toronto, Canada 
}

\begin{document}
\ninept
\maketitle
\begin{abstract}
% It is now known that the expressive power of graph convolutional network (GCN) does not grow infinitely with the number of layers.
% Instead, the GCN output approaches a subspace spanned by the first eigenvector of the normalized graph Laplacian matrix $\widetilde{\L}$ with the convergence rate characterized by the graph spectrum.
A graph convolutional network (GCN) employs a graph filtering kernel tailored for data with irregular structures. 
However, simply stacking more GCN layers does not improve performance; 
instead, the output converges to an uninformative low-dimensional subspace, where the convergence rate is characterized by the graph spectrum---this is the known \textit{over-smoothing} problem in GCN.
% To promote a deeper GCN ar chitecture with sufficient expressiveness, in this paper, we propose the sparse graph learning incorporating spectrum prior to slow down the convergence.
In this paper, we propose a sparse graph learning algorithm incorporating a new spectrum prior to compute a graph topology that circumvents over-smoothing while preserving pairwise correlations inherent in data. 
Specifically, based on a spectral analysis of multilayer GCN output, we derive a spectrum prior for the graph Laplacian matrix $\L$ to robustify the model expressiveness against over-smoothing.
% relate the expressive power of a GCN to a spectrum prior 
Then, we formulate a sparse graph learning problem with the spectrum prior, solved efficiently via block coordinate descent (BCD). 
% Given empirical covariance matrix $\bar{\C}$ computed from observable data, we learn a sparse graph Laplacian matrix $\L$ closest to $\bar{\C}^{-1}$ while optimizing the spectrum prior, with the proposed measure to balance the trade-off between the covariance matrix preservation and the spectrum prior.
Moreover, we optimize the weight parameter trading off the fidelity term with the spectrum prior, based 
on data smoothness on the original graph learned without spectrum manipulation.
The output $\L$ is then normalized for supervised GCN training.
Experiments show that our proposal produced deeper GCNs and higher prediction accuracy for regression and classification tasks compared to competing schemes.
%without explicit spectrum optimization.
\end{abstract}
\section{Introduction}
\label{sec:intro}
Given a defined graph structure, \textit{graph convolutional networks} (GCN)  \cite{kipf2017semi} perform graph filtering and point-wise non-linear operations (\eg, ReLU) in a sequence of neural layers for different tasks, such as graph signal interpolation, denoising, and node classification \cite{kipf2017semi,hgcn_ijcai19,do2020graph}.
However, it has been observed that node representations become indistinguishable (known as \textit{over-smoothing}) and prediction performance quickly degrades as the number of layers grows \cite{li2018deeper,li2019deepgcns}.
This undesirable phenomenon limits GCN's ability to learn appropriate representations from high-order neighborhood and motivates recent research to alleviate the over-smoothing problem \cite{chen2020simple,pei2020geom,oono2020graph,rong2020dropedge}. 
%and has been analyzed theoretically in \cite{oono2020graph}.

Existing works can be classified into two categories depending on whether the graph topology is modified. 
% modification to the graph topology.
The first category of methods focus on novel network architecture designs given a \textit{fixed} graph, \textit{e.g.}, using skip connections to combine features with various receptive fields \cite{xu2018representation}, 
%decoupled transformation and propagation to enable adaptive receptive field \cite{liu2020towards}, 
residual link and identity mapping to enable generalized graph filtering \cite{chen2020simple}, and geometric aggregation to capture long-range dependencies in the graph \cite{pei2020geom}. 
%These approaches are based on the assumption that the graph is given and the modification of graph topology is not considered. 
%\red{this sounds contradictory. u said there are two categories based on modification of graph topology. then u say these approaches do not consider modification of graph topology? do u mean instead given a known graph topology, then modify?} \blue{(revised) the first category does not modify graph topology, while the second one does.}

On the other hand, the second category of methods stress the importance of graph choice in alleviating over-smoothing.
\cite{oono2020graph} theoretically showed that the GCN output approaches an invariant subspace $\cM$ spanned by the first eigenvector of the normalized graph Laplacian matrix $\tilde{\L}$---the subspace is uninformative beyond the number of connected components and node degrees.
Convergence rate of the distance $d_{\cM}$ between the GCN output and $\cM$ 
is characterized by the \textit{graph spectrum} of $\bar{\L}$, \textit{i.e.}, $\bar{\L}$'s eigenvalues determined by the graph topology.

To slow down convergence, \cite{oono2020graph} analyzed an upper bound for $d_{\cM}$ determined by ``eigen-gap''\footnote{The relationship between convergence rate and eigen-gap of a matrix is found also in Perron-Frobenius theorem for a discrete-time Markov chain \cite{meyer10} and the power iteration method in numerical linear algebra \cite{golub12}.}: the difference between the first and second \textit{dominant eigenvalues}\footnote{A dominant eigenvalue is the largest eigenvalue in magnitude.} of matrix $\P = \I - \tilde{\L}$---typically the first two eigenvalues of $\tilde{\L}$.
Given that sparser graphs in general have smaller eigen-gaps (\eg, a complete unweighted graph has the maximum eigen-gap  $2$), 
\cite{oono2020graph} showed that for random Erd\"{o}-R\'{e}nyi graphs with edge probability $p$, sparser graphs (smaller $p$) converge to the aforementioned subspace at a slower pace. 
Similar in approach to achieve graph sparsity, \cite{rong2020dropedge} randomly removed edges from a pre-chosen graph topology in layers during training, resulting in more expressiveness in the trained GCNs.
However, \textit{these methods implicitly optimized %graph spectra 
eigen-gaps
by sparsifying graphs heuristically and may remove strong correlation edges that are essential for effective graph filtering, resulting in sub-optimal GCN performance.} 
%\red{by optimizing graph spectra, I assum u still mean here minimizing eigen-gap.}
%\blue{(revised) yes, for \cite{oono2020graph,rong2020dropedge}, they are minimizing eigen-gap}

% Orthogonally, in the field of \textit{graph signal processing} (GSP), selecting the graph to capture the underlying pairwise node correlations is a fundamental procedure.
% % Orthogonally, \textit{graph signal processing} (GSP) studies discrete signals residing on combinatorial graphs \cite{ortega18ieee,cheung18}. 
% % A key assumption in GSP is that an underlying \textit{similarity graph} capturing pairwise correlations is available as input, before spectral filters are designed and applied to signals on top based on spectral graph theory \cite{chung96}.
% Specifically, given a training set of graph signal observations generated from the same statistical model, there exist many graph learning algorithms \cite{articleglasso,Banerjee7,egilmez17} that compute a most likely sparse inverse covariance matrix (interpreted as a generalized graph Laplacian matrix), which is subsequently used for tasks like compression \cite{hu15,hu15spl}, denoising \cite{pang17,dinesh20} and interpolation \cite{chen21}. 

In contrast, in this paper we propose a sparse graph learning algorithm incorporating a new spectrum prior to mitigate over-smoothing while preserving pairwise correlations inherent in data, resulting in deeper and more expressive GCNs. 
% Unlike \cite{oono2020graph,rong2020dropedge} without explicit spectrum optimization, in this paper, \textit{we directly optimize the graph spectrum with the proposed sparse graph learning algorithm}.
Specifically, inspired by the spectral analysis of multilayer GCN in \cite{oono2020graph}, \textit{we derive a new spectrum prior for $\L$ to robustify the model expressiveness against over-smoothing}. 
%by assuming a \textit{Gaussian Markov Random Field} (GMRF) model \cite{rue2005gaussian} for signals characterized by $\L$.
Given empirical covariance matrix $\bar{\C}$ computed from observable data, we formulate a sparse graph learning problem combining the graphical lasso (GLASSO) objective \cite{articleglasso} with the new spectrum prior, which can be solved efficiently using a variant of \textit{block coordinate descent} (BCD) \cite{wright2015coordinate}.
Moreover, we optimize the weight parameter trading off the GLASSO objective with the spectrum prior, based on observable data smoothness with respect to (w.r.t.) the original graph learned without spectrum manipulation.

Compared with competing schemes \cite{oono2020graph,rong2020dropedge}, by directly optimizing the spectrum we avoid random dropping of strong correlation edges, and thus enhance prediction accuracy for regression and classification tasks. %\red{prediction of what? is GCN always a prediction task?} \blue{(revised)}. 
Moreover, the designed spectrum prior considers the overall eigenvalue distribution rather than the eigen-gap alone, which is more effective in preserving GCN expressiveness.
Different from graph learning algorithms in \cite{articleglasso,Banerjee7,egilmez17,hu15,chen21} that compute a most likely sparse inverse covariance matrix (interpreted as a generalized graph Laplacian matrix), we additionally incorporate the spectrum prior to combat over-smoothing towards deeper GCNs. 

The learned graph is normalized and used for supervised GCN training. 
Experiments show that our proposal produced deeper GCN models with improved performance compared to existing schemes \cite{chen2020simple,pei2020geom,oono2020graph,rong2020dropedge}.
We summarize our contributions as follows.
\begin{enumerate}
\item We design a new spectrum prior 
% as a linear constraint 
for graph Laplacian $\L$ to robustify GCN expressiveness against over-smoothing based on a spectral analysis of multilayer GCN output.
\item We formulate a sparse graph learning problem incorporating the proposed spectrum prior, solved efficiently to preserve pairwise correlation while promoting a desirable spectrum.
\item We optimize the weight parameter, trading off the GLASSO objective with the new spectrum prior, for optimal performance in different learning tasks.
%\item 
% The learnt graph is used for supervised training of GCN, which is validated experimentally to induce deeper GCN model and higher prediction accuracy. 
%\red{is this a contribution? or just experimentally testing your hypothesis and algorithm?} \blue{(revised) moved to the above paragraph}
\end{enumerate}

% \section{Related Works}
% \label{sec:related}
% \input{related.tex}

\section{Preliminaries}
\label{sec:pre}
\subsection{Notations}
An undirected weighted graph $\cG(\cV,\cE,\W)$ is defined by a set of $N$ nodes $\cV = \{1, \ldots, N\}$, edges $\cE = \{(i,j)\}$, and a symmetric \textit{adjacency matrix} $\W$. 
$W_{i,j} \in \mathbb{R}$ is the edge weight if $(i,j) \in \cE$, and $W_{i,j} = 0$ otherwise. 
Self-loops may exist, in which case $W_{i,i} \in \mathbb{R}$ is the weight of the self-loop for node $i$.
Diagonal \textit{degree matrix} $\D$ has diagonal entries $D_{i,i} = \sum_{j} W_{i,j}, \forall i$. 
A \textit{combinatorial graph Laplacian matrix} $\L$ is defined as $\L \triangleq \D - \W$, which is \textit{positive semi-definite} (PSD) for a positive graph \cite{cheung18}. 
If self-loops exist, then the \textit{generalized graph Laplacian matrix} $\cL$, defined as $\cL \triangleq \D - \W + \text{diag}(\W)$, is typically used.

\subsection{Vanilla GCN}
\label{sec:vanilla_gcn}

For a given graph $\cG(\cV,\cE,\W)$, a GCN \cite{kipf2017semi} associated with $\cG$ is defined as follows.
Denote by $\widetilde{\W} \triangleq \W + \I$ and $\widetilde{\D} \triangleq \D + \I$ the adjacency and degree matrices augmented with self-loops, respectively. 
The augmented normalized Laplacian \cite{wu19simpgcn} is defined by $\widetilde{\L} \triangleq \I - \widetilde{\D}^{-1/2} \widetilde{\W} \widetilde{\D}^{-1/2}$, and we set $\P \triangleq \I - \widetilde{\L}$.
Let $L,C \in \mathbb{N}^+$ be the layer and channel sizes, respectively. 
With weights $\mathbf{\Theta}^{(l)} \in \mathbb{R}^{C \times C}$, $l \in \{1,\ldots,L\}$, the GCN is defined by $f = f_L \circ \dots \circ f_1$ where $f_l: \mathbb{R}^{N \times C} \mapsto \mathbb{R}^{N \times C}$ is defined by $f_l(\X) \triangleq \sigma( \P \X \mathbf{\Theta}^{(l)})$, where $\sigma$ denotes the nonlinear activation operator $\operatorname{ReLU}$.

\section{Spectral Analysis}
\label{sec:analyze}
Based on the spectral analysis of multilayer GCN output in \cite{oono2020graph}, we discuss the motivation of sparse graph learning with a spectrum prior to alleviate over-smoothing and induce deeper GCNs.
First, we show that the convergence of GCN output to a low-dimensional invariant subspace is characterized by the graph spectrum. 
To robustify model expressiveness, we propose a linear spectrum prior,  which will be incorporated into a sparse graph learning algorithm in the sequel.

\subsection{Oversmoothing in Multilayer GCN}
% GCN 
As defined in Sec.\ref{sec:vanilla_gcn}, for a multilayer GCN model associated with $\cG$, each layer $f_l(\X) \triangleq \sigma( \P \X \mathbf{\Theta}^{(l)})$ consists of three basic operators: the graph operator $\P$, the filter $\mathbf{\Theta}^{(l)}$, and the activation $\sigma$. 
As proved in \cite{oono2020graph}, each of the three operators leads to a decrease of the distance between the output of $l$-th layer $\X^{(l)}$ and the invariant subspace $\cM$. 
Here, we focus on the graph operator $\P$, which is determined by the graph topology.

Specifically, denote by $\{ \v_1, ..., \v_N \}$ the orthonormal eigenvectors of $\P$ corresponding to eigenvalues $\lambda_1 \leq \dots \leq \lambda_N$.
Suppose $\cG$ has $M$ connected components. 
Then, we have $\lambda \triangleq \max_{n = 1,\ldots,N-M} |\lambda_n| < 1$ and $\lambda_{N-M+1} = \dots = \lambda_N = 1$.
We can then uniquely write $\X \in \mathbb{R}^{N \times C}$ as $\X = \Sigma_{n=1}^N \v_n \otimes \w_n$,
%$\red{\sum_{n=1}^N w_n \v_n \v_n^\top}$
where $\w_1,\dots,\w_n \in \mathbb{R}^{ C}$ are the coefficients w.r.t. the basis $\{ \v_1, ..., \v_N \}$, and $\otimes$ is the Kronecker product. 
% The distance to the invariance space $\cM$ is given as,
% \begin{align}
%     d_{\cM}^2(\X) = \Sigma_{n=1}^{N-M} || \w_n ||^2.
% \end{align}
% \begin{align}
%     \P \X & = \Sigma_{n=1}^N \P \v_n \otimes \w_n \\
%           & = \Sigma_{n=1}^{N-M} \P \v_n \otimes \w_n + \Sigma_{n=N-M+1}^N \P \v_n \otimes \w_n \\
%           & = \Sigma_{n=1}^{N-M} \v_n \otimes (\lambda_n \w_n) + \Sigma_{n=N-M+1}^N \P \v_n \otimes \w_n.
% \end{align}
% Since the eigenspace spanned by $\{ \v_{N-M+1}, ..., \v_N \}$ is invariant under $\P$, for $n \in [N-M+1, N]$, $\P \v_n$ can be expressed as a linear combination of $\v_m$ ($m \in [N-M+1, N]$). 
% \begin{align}
%   d_{\cM}(\P \X) \leq \lambda d_{\cM}(\X), \\
%   d_{\cM}(\X \mathbf{\Theta}) \leq s d_{\cM}(\X), \\
%   d_{\cM}( \sigma (\X) ) \leq d_{\cM}(\X),
% \end{align}
% where $s:= \sup_{l=1,\ldots L} s_l$ and $s_l$ is the maximum singular value of $\mathbf{\Theta}^{(l)}$.
When applying the operator $\P$ to $\X$, the distance to the invariance space $\cM$ is given by
\begin{align}
    d_{\cM}^2(\P \X) &= \Sigma_{n=1}^{N-M} || \lambda_n \w_n ||^2 \label{eq:d_PM0} \\
    &\leq \Sigma_{n=1}^{N-M} \lambda^2 || \w_n ||^2 
    = \lambda^2 d_{\cM}^2(\X), \label{eq:d_PM}
\end{align}
which shows that the graph operator $\P$ reduces the distance to the invariance space $\cM$ at a rate characterized by the graph spectrum $\lambda_n$ for $n \in \{1, \dots, N-M\}$.

\subsection{Spectrum Prior to Alleviate Over-smoothing}

For given coefficients $\{\w_1,\dots,\w_{N-M}\}$, 
%determined by the eigen-basis of $\P$ \red{weights are independent of the eigenvectors, right?} \blue{no, $\w_n$ is the coefficient of $\X$ wrt to the eigen-basis of $\P$, should be determined by eigenvectors}, 
to slow down the convergence to $\cM$, the eigenvalues should be optimized as
\begin{equation} \label{eq:obj_max}
\max_{\lambda_1, \ldots, \lambda_{N-M}} \Sigma_{n=1}^{N-M} \lambda_n^2 || \w_n ||^2
, 
\mbox{s.t.} ~~ -1 < \lambda_1 \leq \dots \leq \lambda_{N-M} < 1,
\end{equation}
%\red{should not be an unconstrained optimization problem, right? otherwise  I can set $\lambda_n$ to $\infty$?} \blue{(revised) should consider range of $\lambda$}
% To relief over-smoothing, the new graph operator $\P'$ corresponding to the learned graph must satisfy $d_{\cM}(\P' \X) > d_{\cM}(\P \X)$.
% Without any prior knowledge of $\w_n$, this requires each eigenvalue $\lambda'_n$ of $\P'$ to satisfy $|\lambda'_n| > |\lambda_n|$ for $n \in \{1, \dots, N-M\}$---this spectral constraint is difficult to maintain during optimization. 
%\red{why?} \blue{(revised) this is to avoid optimization in spectral domain which is slow when graph is large, like in Saghar's ICASSP paper}.
where the objective depends on the spectrum of $\P$ and the coefficient $\w_n$ of feature $\X$. 
To derive a spectrum prior as a function of $\L$, we relate $\lambda_n$ and $\w_n$ to the spectrum property of $\L$ as follows. 

First, $\P$ is obtained from $\L$ via normalization to ensure $\P$ has eigenvalues in the range $[-1,1]$.
Instead of using the normalization in \cite{kipf2017semi}, we adopt the following procedure to derive a linear spectrum prior for $\L$. 
Let $0 \leq \mu_1 \leq \dots \leq \mu_N$ be the eigenvalues of $\L$.
$\mu_1 \I_N$ is subtracted from $\L$, \textit{i.e.}, $\L_0 = \L - \mu_1 \I_N$, so that the smallest eigenvalue of $\L_0$ is $0$, and correspondingly the largest eigenvalue of $\P$ is $1$.
% Then, the component of $\X$ in $\cM$ does not vanish after applying $\P$ operator. \red{I don't get this.}
%\red{not clear} \blue{if the largest eigenvalue of $\P$ is smaller than 1, then $\P\X$ will reduce the corresponding frequency component, which is the component in the invariant space} 
Then, $\L_0$ is scaled as  
% $\L_{norm} = \frac{2}{\mu_{max}} \L_0$, $\P = \I - \L_{norm}$,
\begin{equation} \label{eq:L_norm}
    \L_{\textrm{norm}} = \frac{2}{\mu_{\textrm{max}}} \L_0 , \quad \P = \I - \L_{\textrm{norm}},
\end{equation}
where $\mu_{\textrm{max}} > \mu_N$ is set to ensure that the eigenvalues of $\L_{\textrm{norm}}$ are in the range $[0, 2]$. 
Thus,
\begin{equation}
    \lambda_n = 1 - 2(\mu_{N-n+1} - \mu_1)/\mu_{\textrm{max}},
    %, ~~\w_n = \v_{N-n+1},
\end{equation}
where $\mu_{N-n+1}$ has index $N-n+1$ because the eigenvalues of $\P$ and $\L$ have reverse orders. 
Moreover, from the procedure above we can see $\P$ and $\L$ share the same eigen-basis. 
Thus, the coefficient $\u_n$ of $\X$ w.r.t. the eigen-basis of $\L$ is given as $\w_n = \u_{N-n+1}$.

Next, to examine $\u_n$, we assume that the model of $\X$ is a \textit{Gaussian Markov Random Field} (GMRF) \cite{rue2005gaussian} w.r.t. $\cG$, with covariance matrix $\mathbf{\Sigma}^{-1} = \L + \delta \I$, where $1/\delta$ is interpreted as the variance of the DC component for $\X$ \cite{gadde2015probabilistic}.
The expected energy of $\| \u_n \|^2$ 
%\red{$\w_n$ are coefficients of what?} \blue{changed to $\v_n$, coefficient of $\X$ wrt to the eigen-basis of $\L$. same as $\w_n$ since eigen-basis of $\L$ and $\P$ is the same because the normalization is just a scaling}
is $\text{E}[ \| \u_n \|^2 ] = 1/(\delta + \mu_n)$ \cite{zeng2017bipartite}.
% $\text{E}[ \| \w_n \|^2 ] = \frac{1}{\delta + \mu_{N-n+1}}$,
% \begin{equation} \label{eq:signal}
%     E[ || \w_n ||^2 ] = \frac{1}{\delta + \mu_{N-n+1}},
% \end{equation}
%where $\mu_{N-n+1}$ has the index $(N-n+1)$ because the eigenvalues of $\P$ and $\L$ have opposite orders. 
% The objective function \eqref{eq:obj_max} becomes
% \begin{equation} \label{eq:obj_max2}
%     \max_{\lambda_1, ..., \lambda_{N-M}} \Sigma_{n=1}^{N-M} \frac{\lambda_n^2}{\delta+ \mu_{N-n+1}}.
% \end{equation}
With $\delta = 0$, the objective \eqref{eq:obj_max} becomes 
\begin{equation} 
    \max_{\mu_{M+1}, ..., \mu_{N}} \Sigma_{n=M+1}^{N} (1 - \frac{2}{\mu_{\textrm{max}}} (\mu_n - \mu_1) )^2 / \mu_n,
\end{equation}
which can be further simplified to
\begin{equation}
    \max_{\mu_{M+1}, ..., \mu_{N}}  \Sigma_{n=M+1}^{N} (1 + \frac{2\mu_1}{\mu_{\textrm{max}}} )^2 / \mu_n + (\frac{2}{\mu_{\textrm{max}}})^2 \mu_n 
\end{equation}
and the objective function decreases monotonically when $\mu_n < \frac{\mu_{\textrm{max}} + 2\mu_1}{2}$. By setting $\mu_{\textrm{max}} \geq 2 (\mu_N - \mu_1)$, the objective becomes $\min \Sigma_{n=M+1}^{N} \mu_n$. Since $\mu_{1}, \ldots, \mu_{M}=0$, the objective is further simplified to $\min \| \boldsymbol{\mu} \|_1$, where $\boldsymbol{\mu} = [\mu_{1}, \dots, \mu_{N}]$.
% \begin{equation}
%     \min_{\mu_{M+1}, ..., \mu_{N}}  \quad \Sigma_{n=M+1}^{N} \mu_n = | \boldsymbol{\mu} |_1,
% \end{equation}

In summary, assuming the model of $\X$ is a GMRF specified by $\L$ with eigenvalues $0 \leq \mu_1 \leq \dots \leq \mu_N$, $\boldsymbol{\mu} = [\mu_{1}, \dots, \mu_{N}]$, convergence to the invariant space $\cM$ is slowed down via
\begin{equation} \label{eq:l1_mu}
    \min_{\L}  \quad \| \boldsymbol{\mu} \|_1,
\end{equation}
where $\mu_{\textrm{max}} \geq 2 (\mu_N - \mu_1)$ for $\L$ normalization to produce $\P$.

% \subsection{Discussion} 
% From \eqref{eq:l1_mu}, we can see the spectrum optimization is determined by the graph topology, indicating that the oversmoothing can be alleviated via graph learning. 
% Moreover, the adopted normalization and signal prior assumption lead to a linear spectrum prior for $\L$ which enable efficient graph learning algorithm as discussed in the next section.

\vspace{0.05in}
\noindent
\textbf{Relation to Weight Scaling Scheme in \cite{oono2020graph}}
Based on the upper-bound of $d_{\cM}^2(\P \X)$ in (\ref{eq:d_PM}), \cite{oono2020graph} proved that, with initial value $\X^0$, $d_{\cM}(\X^{(l)})$ satisfies $d_{\cM}(\X^{(l)}) \leq (s \lambda)^l d_{\cM}(\X^0)$,
% \begin{equation}
%   d_{\cM}(\X^{(l)}) \leq (s \lambda)^l d_{\cM}(\X^0),
% \end{equation}
where $s:= \sup_{l=1,\ldots L} s_l$ and $s_l$ is the maximum singular value of $\mathbf{\Theta}^{(l)}$.
In particular, $d_{\cM}(\X^{(l)})$ exponentially converges to 0 if $s \lambda < 1$.
Hence, \cite{oono2020graph} proposed to normalize the weight $\mathbf{\Theta}^{(l)}$ so that $s_l \lambda > 1$ in order to slow down the convergence to the invariant subspace.

However, since the eigenvalues of $\P$ are generally different from $\lambda$, the upper bound of $d_{\cM}^2(\P \X)$ in (\ref{eq:d_PM}) is so loose that the weight scaling scheme proposed in \cite{oono2020graph} has limited effect in avoiding over-smoothing.
In contrast, our proposed spectrum prior in \eqref{eq:l1_mu} considers the entire spectrum instead of only the second largest $\lambda$, leading to improved performance as validated in our experiments.

\section{Sparse Graph Learning with Spectrum Prior for GCN Training}
\label{sec:algo}
In this section, we propose a new sparse graph learning algorithm using the proposed graph spectrum prior. 
Further, we design a measure to optimally trade off pairwise correlation preservation with the spectrum prior, based on smoothness of observable data on the original graph learned without spectrum manipulation.

\subsection{Problem Formulation}

We incorporate the spectrum prior in \eqref{eq:l1_mu} into the GLASSO formulation \cite{articleglasso}, resulting in the following graph learning objective:
% \begin{align}
% & \text{Tr}(\mathbf{L \bar{\C}} ) -\log \det \mathbf{L}+ \rho \; \| \mathbf{L} \|_1 + \sigma | \boldsymbol{\mu} |_1 \\
% =  ~~ & \text{Tr}(\mathbf{L \bar{\C}} ) -\log \det \mathbf{L}+ \rho \; \| \mathbf{L} \|_1 + \sigma \text{Tr}(\mathbf{L}) \\
% =  ~~ & \text{Tr}( \mathbf{L}  (\bar{\C}  + \sigma \mathbf{I})) -\log \det \mathbf{L}+ \rho \; \| \mathbf{L} \|_1  \label{eq:new_obj} 
% \end{align}
% The spectral graph learning is formulated as
\begin{align}
% & \min_{\mathbf{L} \succeq \mathbf{0} } ~~ \text{Tr}( \mathbf{L}  (\bar{\C}  + \sigma \mathbf{I})) -\log \det \mathbf{L}+ \rho \; \| \mathbf{L} \|_1,
& \min_{\mathbf{L} \succeq \mathbf{0} } ~~ \text{Tr}(\mathbf{L \bar{\C}} ) -\log \det \mathbf{L}+ \rho \; \| \mathbf{L} \|_1 + \sigma \text{Tr}(\mathbf{L}),
\label{eq:obj} 
\end{align}
where the spectrum prior is $\text{Tr}(\mathbf{L}) = \| \boldsymbol{\mu} \|_1$. $\bar{\C}$ is the input empirical covariance matrix computed from observable data, $\rho > 0$ is a shrinkage parameter for the $\ell_1$-norm of $\L$, and $\sigma$ is the weight for spectrum prior. Next, we discuss the computation of $\sigma$ to trade off the GLASSO objective with the spectrum prior.

\subsection{Computing Tradeoff Parameter $\sigma$}

Based on the smoothness of the observable data $\F \in \mathbb{R}^{N \times K}$ on the original graph learned 
given $\bar{\C}$ without spectrum manipulation, we determine weight $\sigma$ in \eqref{eq:obj} to trade off preservation of pairwise correlation inherent in $\bar{\C}$ with alleviation of over-smoothing. 
The idea is the following: 
if data $\F$ is smooth w.r.t. the original GLASSO output $\hat{\L}$ without spectrum prior ($\sigma=0$), then $\F$ has energy mostly in the invariant subspace $\cM_0$ of $\hat{\L}$ spanned by the eigenvector of the lowest frequency.
That means convergence to $\cM_0$ has little impact on prediction accuracy, and hence the spectrum prior can be removed, \textit{i.e.}, $\sigma=0$. 
Otherwise, the spectrum prior should be assigned higher weight to slow down convergence.
% In other words, the mismatch between $\F$ and the original GLASSO output given $\bar{\C}$ indicates the importance of the spectrum prior, and can be used to compute $\sigma$.

To quantify signal smoothness, given the original GLASSO output $\hat{\L}$, we define the measure using a variant of the quadratic \textit{graph Laplacian regularizer} (GLR) \cite{pang17}, \ie, 
\begin{equation}
    \mathrm{M}_{\hat{\L}}(\F)  = \frac{\Tr (\F^{\top} \hat{\L} \F )}{\hat{\mu}_N \Tr(\F^{\top}  \F)},
\end{equation}
which measures the smoothness of the data w.r.t. graph specified by $\hat{\L}$, normalized by the signal energy. 
$\hat{\mu}_N$ is the largest eigenvalue of $\hat{\L}$, used here to normalize $\mathrm{M}_{\hat{\L}}(\F)$ to the range $[0,1]$.

Weight $\sigma$ should increase monotonically with $\mathrm{M}_{\hat{\L}}(\F)$ in $[0,1]$, and thus we set $\sigma$ to be a scaled and shifted logit function \cite{cramer2003origins}:
\begin{equation} \label{eq:sigma}
\sigma =  \ln \left( \frac{1 + \mathrm{M}_{\hat{\L}}(\F)}{1 - \mathrm{M}_{\hat{\L}}(\F)} \right).
\end{equation} 
We see that when $\mathrm{M}_{\hat{\L}}(\F)$ is small, $\sigma$ is small, and when $\mathrm{M}_{\hat{\L}}(\F)$ approaches $1$, $\sigma$ approaches infinity.

\subsection{Algorithm Design}
\label{sec:bcd}

Given computed $\sigma$ and the objective in \eqref{eq:obj}, we design an algorithm as summarized in Algorithm\;\ref{algo:spectral_GL}. 
We call the algorithm \textit{Sparse Graph Learning with Spectrum Prior for GCN}, named \textbf{SGL-GCN}.

By combining $\text{Tr}(\mathbf{L \bar{\C}} )$ and $\sigma \text{Tr}(\mathbf{L})$ as $\text{Tr}(\mathbf{L (\bar{\C}+ \sigma \I)} )$,  we solve \eqref{eq:obj} iteratively using a variant of the \textit{block coordinate descent} (BCD) algorithm \cite{wright2015coordinate}. 
Specifically, similarly done in \cite{bagheri21icassp}, we solve the \textit{dual} of GLASSO as follows.
First, note that the $\ell_1$-norm in \eqref{eq:obj} can be written as 
\vspace{-0.05in}
\begin{align}
\|\L\|_1 = \max_{\| \U \|_\infty\leq 1} ~~\text{Tr}(\L\U) 
\end{align}
where $\| \U \|_\infty$ is the maximum absolute value element of the symmetric matrix $\U$. 
The dual problem of GLASSO that seeks an estimated covriance matrix $\C = \L^{-1}$ can now be written as
%Denote by $\C = \P^{-1}$. 
\begin{align}
\min_{\C } ~~ -\log \det \C, 
~~~\mbox{s.t.} ~~ 
\| \C - (\bar{\C}+ \sigma \I) \|_\infty \leq \rho.
\label{eq:glasso_dual}    
\end{align}
% where $\C =\bar{\C}+\U$  implies that the primal and dual variables are related via $\L = {(\bar{\C}+\U)}^{-1} $ \cite{Banerjee7}.
To solve \eqref{eq:glasso_dual}, 
we update one row-column pair of $\C$ in \eqref{eq:glasso_dual} in each iteration following optimization procedure in \cite{Banerjee7}.

In summary, our algorithm to solve \eqref{eq:obj} is as follows.
We minimize the GLASSO terms in \eqref{eq:obj} by solving its dual \eqref{eq:glasso_dual}---iteratively updating one row / column of $\C$ at a time.
We repeat these two steps till convergence.
Note that both steps are computed using covariance $\C$ directly, and thus inversion to graph Laplacian $\L = \C^{-1}$ is not necessary until convergence, when we output a solution.

\begin{algorithm}[t]
\caption{Sparse Graph Learning with Spectrum Prior for GCN}
\label{algo:spectral_GL}
\begin{small}
\begin{algorithmic}[1]
\Require Empirical covariance matrix $\bar{\C}$, observable data $\F$
\Ensure Graph operator $\P$ for GCN model
\State Obtain the original GLASSO output $\hat{\L}$ given $\bar{\C}$ with $\sigma = 0$ via graph learning algorithm described in Sec\,\ref{sec:bcd}.
\State Compute the weight value $\sigma$ given $\hat{\L}$ and $\F$.
\State Obtain the sparse graph learning output $\L$ incorporating GLASSO objective and spectrum prior via algorithm in Sec\,\ref{sec:bcd}.
\State Normalize $\L$ with (\ref{eq:L_norm}) to produce $\P$.
\State Use $\P$ for GCN training.
\end{algorithmic}
\end{small}
\end{algorithm}

\section{Experiments}
\label{sec:exp}
We conducted experiments to validate our graph learning proposal that alleviates over-smoothing and improves prediction accuracy by comparing against recent proposals, including DropEdge \cite{rong2020dropedge}, Oono's scheme \cite{oono2020graph}, GCNII \cite{chen2020simple} and Geom-GCN \cite{pei2020geom}.

% \subsection{Dataset and Evaluation Metric} 
\subsection{Dataset and Experiment Settings} 
% \noindent \textbf{Dataset.} 
For regression, we used the \texttt{METR-LA} dataset \cite{li2018dcrnn_traffic} containing traffic speed data in four months (from March 1st 2012 to June 30th 2012) from 207 sensors in the Los Angeles County. 
The sensors sampled the speed data every 5 minutes. 
Our task is to predict the current traffic speed using historical speed data in the past 50 minutes as the input feature. 
We randomly sampled 70\% data for training, 20\% for validation, and 10\% for testing. 
The empirical covariance $\bar{\C}$ was computed using all the observations in the training data.

For node classification, we used \texttt{Cornell}, \texttt{Texas}, and \texttt{Wisconsin} datasets, which are from the \texttt{WebKB} dataset \cite{pei2020geom}. 
These dataset are web networks, where nodes and edges represent web pages and hyperlinks, respectively. The feature of each node is the bag-of-words representation of the corresponding page. 
We followed the experimental setting in \cite{pei2020geom} for node classification. 
$\bar{\C}$ was the inverse of the graph Laplacian constructed using node feature similarity. 
In particular, we constructed K-NN graph ($K=10$), with edge weights computed as $w_{ij}=\exp (-\| \f_i - \f_j\|^2/2\gamma)$ ($\gamma=5$), where $\f_i$ is the feature for node $i$. 

% \noindent \textbf{Evaluation Metric.}
% We use Mean Squared Error (MSE) as our evaluation metric for traffic prediction defined, $\text{MSE}(\mathbf{x},\mathbf{\hat{x}} ) = 1/N\sum_{i=1}^{N}{(x_i - \hat{x}_i)^2 }$,
% % \begin{equation}
% % \text{MSE}(\mathbf{x},\mathbf{\hat{x}} ) = \frac{\sum_{i=1}^{N}{(x_i - \hat{x}_i)^2 }}{N}
% % \end{equation}
% where $\mathbf{x} = [ x_1, \dots, x_N ]$ are the ground truth values, and $\mathbf{\hat{x}} = [\hat{x}_1, \dots, \hat{x}_N]$ are the predicted values. For node classification, we use the classification accuracy.

% \subsection{Experimental Settings}

For graph learning, sparsity parameter $\rho$ in \eqref{eq:obj} was set to $10^{-4}$. 
For normalization of $\L$, $\mu_{\textrm{max}}=11$ for \texttt{METR-LA} and $\mu_{\textrm{max}}=1$ for \texttt{Web-KB}.
GCN training was implemented in PyTorch and optimized by Adam \cite{kingma2014adam} with initial learning rate set to $0.01$ and weight decay set to $5e-5$. 
Our GCN model was consisted of $L$ GCN blocks ($L = [1, 10]$) and two linear layers. 
% Each GCN block contained a graph convolutional layer, a BatchNorm layer and an activation layer (leaky ReLU). 
%We set the random seed as 123 for sampling training dataset and initializing the network.

\subsection{Validation of Weight Computation}
\begin{figure}[h]
\centering
\includegraphics[width=0.75\columnwidth]{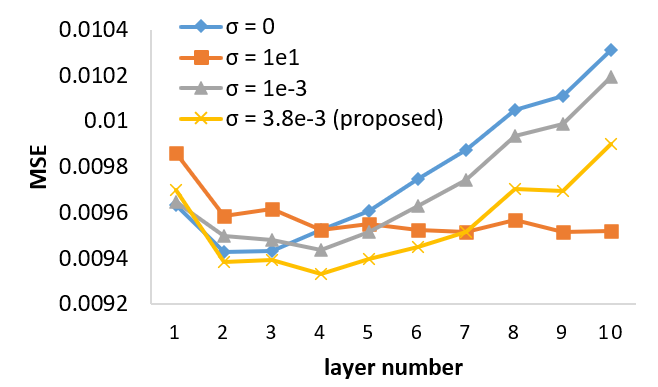}
\vspace{-0.2in}
\caption{MSE results with different layer sizes of GCN models using Laplacian matrices learned with different weighting factors. The proposed scheme balance the convergence rate and the covaraince preservation, achieving the lowest prediction error.}
    \label{fig:sigma}
\end{figure}

To validate the proposed measure to compute $\sigma$, we set $\sigma \in \{0, 1e1, 1e-3 \}$ and compared against the value computed via our proposed scheme, which was $0.0038$ for \texttt{METR-LA} dataset.
Fig.\;\ref{fig:sigma} shows the results for GCN training using graph Laplacian matrices learned using different $\sigma$. 
% Results with 1 layer were omitted since the values were too large.
% As shown in Fig.\;\ref{fig:sigma}, the computed weight balanced preservation of pairwise correlations inherent in $\bar{\C}$ with allevaition of over-smoothing.
Large $\sigma$, \textit{e.g.} $1e1$, mitigated over-smoothing of the GCN model and achieved larger optimal layer number ($7$), but the learned graph deviated too far from $\bar{\C}$ and failed to achieve small MSE.
On the other hand, small $\sigma$, \textit{e.g.} $1e-3$ quickly reduced the MSE with few layers ($2$), but could not reduce MSE with more layers due to over-smoothing.
Meanwhile, our proposed scheme with $\sigma=3.8e-3$ achieved the lowest MSE, indicating the importance of choosing an appropriate weight $\sigma$.
% Thus, it is essential to choose an appropriate weight $\sigma$ to achieve the lowest MSE using our proposed scheme with $\sigma=3.8e-3$, as shown in Fig.\;\ref{fig:sigma}.

% \vspace{-0.05in}

\begin{table}[h]
\centering
\caption{ MSE results ($\times 10^{-3}$) of META-LA dataset with different layer size of GCN models using different schemes. }
\begin{tabular}{l|c|c|c}
\hline
 Methods & 2 layers & 4 layers & 8 layers  \\ \hline \hline
GCN \cite{kipf2017semi} &	10.76	& 11.81 &	17.38 \\ \hline
DropEdge \cite{rong2020dropedge} &	10.79	&12.25&	17.72\\ \hline
Oono's \cite{oono2020graph} &10.79	& 11.89	& 17.32	\\ \hline
% GCNII \cite{chen2020simple} & 10.72	&10.69&	10.96	\\ \hline 
SGL-GCN w/o spectrum & 9.43 & 9.52 & 10.05 \\ \hline
SGL-GCN &	\textbf{9.38}&	\textbf{9.33}&	\textbf{9.70}\\ \hline 

\end{tabular}%
  \label{tab:MSE}%
\end{table}%
\vspace{-0.2in}

\subsection{Comparison with State-of-the-Art Methods}

We compared our method with competing schemes DropEdge \cite{rong2020dropedge} and Oono's scheme \cite{oono2020graph} for traffic prediction task, and additionally included GCNII \cite{chen2020simple} and Geom-GCN \cite{pei2020geom} for node classification. 
The most related schemes are DropEdge and Oono's method which modify graph topologies for GCN training. 
We used the same experimental settings for DropEdge with drop rate $p = 0.3$ and Oono's weight scaling with default $\s_0 = 1$. 
For Geom-GCN, three variants were included using different embedding methods, \textit{i.e.}, Isomap (Geom-GCN-I), Poincare (Geom-GCN-P), and struc2vec (Geom-GCN-S).
For our proposal, $\sigma=0.9661, 0.9147, 0.9147$ for \texttt{Wisconsin}, \texttt{Texas}, and \texttt{Cornell} datasets, respectively.

The resulting MSE of META-LA dataset are shown in Table\;\ref{tab:MSE}, where the optimal results are highlighted in bold font.
We observe that our method had better performance in terms of slowing down over-smoothing and achieving higher prediction accuracy. 
Specifically, DropEdge achieved its best result at the second layer with MSE $0.0107$, while our method increased the optimal layer number to $4$ and achieved lower MSE $0.0093$.
Moreover, by removing the spectrum prior in our proposal, \ie, SGL-GCN w/o spectrum in Table\;\ref{tab:MSE}, the performance was degraded, validating the effectiveness of the spectrum prior.

The test accuracy of WebKB dataset are shown in Table\;\ref{tab:webkb}. 
We selected the performance of the optimal layer number for each scheme. 
We see that our proposal outperformed the state-of-the-art methods in all three datasets. 
Comparing with DropEdge and Oono's, we increased the accuracy by more than $20\%$, which clearly shows the importance of explicit spectrum optimization.

% \begin{figure}[h]
% \centering
% \includegraphics[width=0.8\columnwidth]{fig/dropedge.png}
% \vspace{-0.35in}
% \caption{ MSE results with different layer sizes of GCN models using DropEdge \cite{rong2020dropedge} with different drop rates, compared with our proposal with $\sigma=3.8e-3$.}
%     \label{fig:dropedge}
% \end{figure} 
% \vspace{-15pt}

\begin{table}[h]
\centering
\caption{Test accuracy (\%) of WebKB dataset for different schemes.}
\begin{tabular}{l|c|c|c}
\hline
 Methods & \texttt{Wisconsin} &	\texttt{Texas} & \texttt{Cornell}  \\ \hline \hline
GCN \cite{kipf2017semi} &45.88	&52.16&	52.7  \\ \hline
Dropedge \cite{rong2020dropedge} &	61.62&	57.84&	50.2\\ \hline
Oono's \cite{oono2020graph} &	53.92 &	58.92 &	61.08 \\ \hline
Geom-GCN-I \cite{pei2020geom}&	58.24&	57.58&	56.76 \\ \hline
Geom-GCN-P \cite{pei2020geom}&	64.12&	67.57&	60.81 \\ \hline
Geom-GCN-S \cite{pei2020geom}&	56.67&	59.73&	55.68 \\ \hline
GCNII \cite{chen2020simple} &81.57&	77.84&	76.49	\\ \hline 
SGL-GCN w/o spectrum	& 82.35 & 78.11	& 80.00	\\ \hline
SGL-GCN &	\textbf{85.69}&	\textbf{82.70} &\textbf{82.97}	\\ \hline 
\end{tabular}%
  \label{tab:webkb}%
\end{table}%
\vspace{-0.2in}

\section{Conclusion}
\label{sec:con}
We propose a sparse graph learning algorithm with a new spectrum prior to alleviate over-smoothing problem of GCN model while preserving pairwise correlation inherent in data.
Specifically, a new spectrum prior is designed to robustify the GCN expressiveness against over-smoothing, which is combined with the GLASSO objective for efficient sparse graph learning. The trade-off between the fidelity term and spectrum prior is balanced with the proposed measure that quantifies the data smoothness on the graph learned without spectrum manipulation.
% The spectral graph learning produces the graph Laplacian that slows down the oversmoothing and leads to more optimal depth layers (more expressiveness) and lower loss function values during supervised GCN training.
Compared to competing schemes, our proposal produced deeper GCNs with improved performance.

\vfill\pagebreak

\bibliographystyle{IEEEbib}
\bibliography{strings,refs_v2}

\end{document}